# A Vehicle-Integrated Approach to Digital Twin Deployment for Bridges Through Drive-By Sensing

**Zihao Liu[1], Daigo Kawabe[2], Jiaji Wang[3], Chul-Woo Kim[4] and Mehrisadat Makki Alamdari[1*]**

[1]School of Civil and Environmental Engineering, University of New South Wales, Sydney, Australia
[2]KyoCenseo Inc., Kyoto, Japan
[3]Department of Civil Engineering, The University of Hong Kong, Hong Kong, China
[4]Department of Civil & Earth Resources Engineering, Kyoto University, Kyoto, Japan
*m.makkialamdari@unsw.edu.au

## 1. INTRODUCTION

Ageing and deteriorating bridge infrastructure is a growing challenge globally. Conventional assessment relies on periodic visual inspections and permanently installed structural health monitoring systems: visual inspections are subjective and can miss early-stage deterioration, while fixed sensor networks are costly to install and maintain, making them difficult to deploy across large bridge inventories, particularly in regional areas. Drive-by, or indirect, inspection promises a scalable middle path. An instrumented vehicle records its own vibration while it drives across a bridge. Because the vehicle and the bridge form a coupled dynamic system, known as vehicle-bridge interaction (VBI), the vehicle response carries the signatures of both subsystems [1]. Figure 1 illustrates the concept. Key challenges remain unresolved, including separating the bridge contribution from road roughness and vehicle dynamics, detecting damage under normal traffic, and generalising across bridge types.

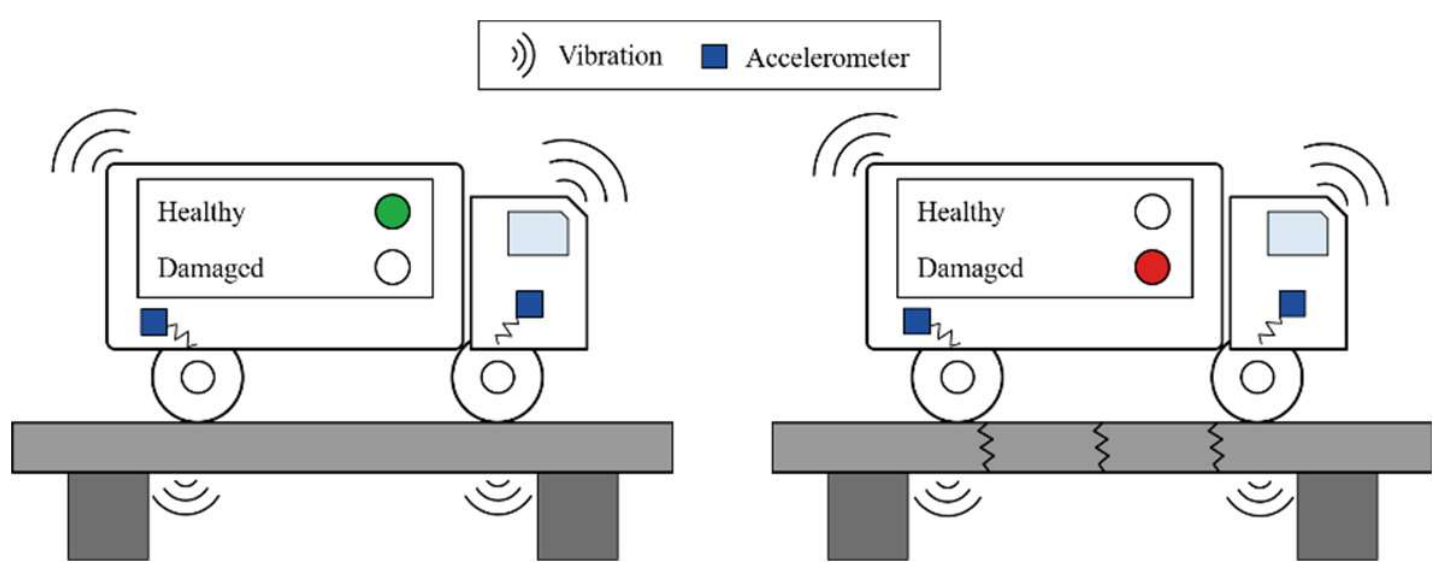

**Figure 1.** Illustration of indirect drive-by inspection.

Drive-by measurements become useful only once they are converted into bridge condition, and this conversion, an inverse problem over the coupled vehicle-bridge system, is where these challenges concentrate. The approach taken in this paper is to create a digital twin of the bridge: a computational replica, updated with every crossing, that embeds the bridge-specific physics the inversion needs. The twin performs two complementary tasks: predicting the response that a hypothesised structural condition would produce, and inferring the condition from the measured response. Both must run at the rate at which crossings occur. Full-order finite-element (FE) simulation of VBI is too slow for this purpose; in the present study, a single simulated crossing of the truss model required about one minute of computation. Neural operators, which learn mappings between function spaces rather than between fixed-dimensional vectors [2], achieve inference times of the order of milliseconds [3] and have recently been applied to structural simulation and bridge health monitoring [3-5]. This paper presents a neural operator digital twin of a real truss bridge with member-level damage, and reports its forward and inverse performance in the drive-by configuration, in which the crossing vehicle's response is the measurement.

## 2. THE ADA BRIDGE DIGITAL TWIN

The testbed is the Old Ada Bridge in Nara Prefecture, Japan, shown in Figure 2: a simply supported steel truss with a main span of 59.2 m and a width of 3.6 m, constructed in 1959 and decommissioned in 2012 [6]. Prior to its removal, ambient and vehicle-induced vibration data were collected on the bridge under 5 artificial damage scenarios, and this benchmark dataset is available from the authors' previous work [6]. The digital twin itself is the neural operator surrogate described below. Its training data are generated by an FE model of the truss, whose 29 members fall into 4 groups (bottom chord, top chord, verticals and diagonals). In each simulated crossing, a two-axle vehicle crosses the deck at constant speed on a fixed road profile, so that differences between crossings are attributable to the structural state alone.

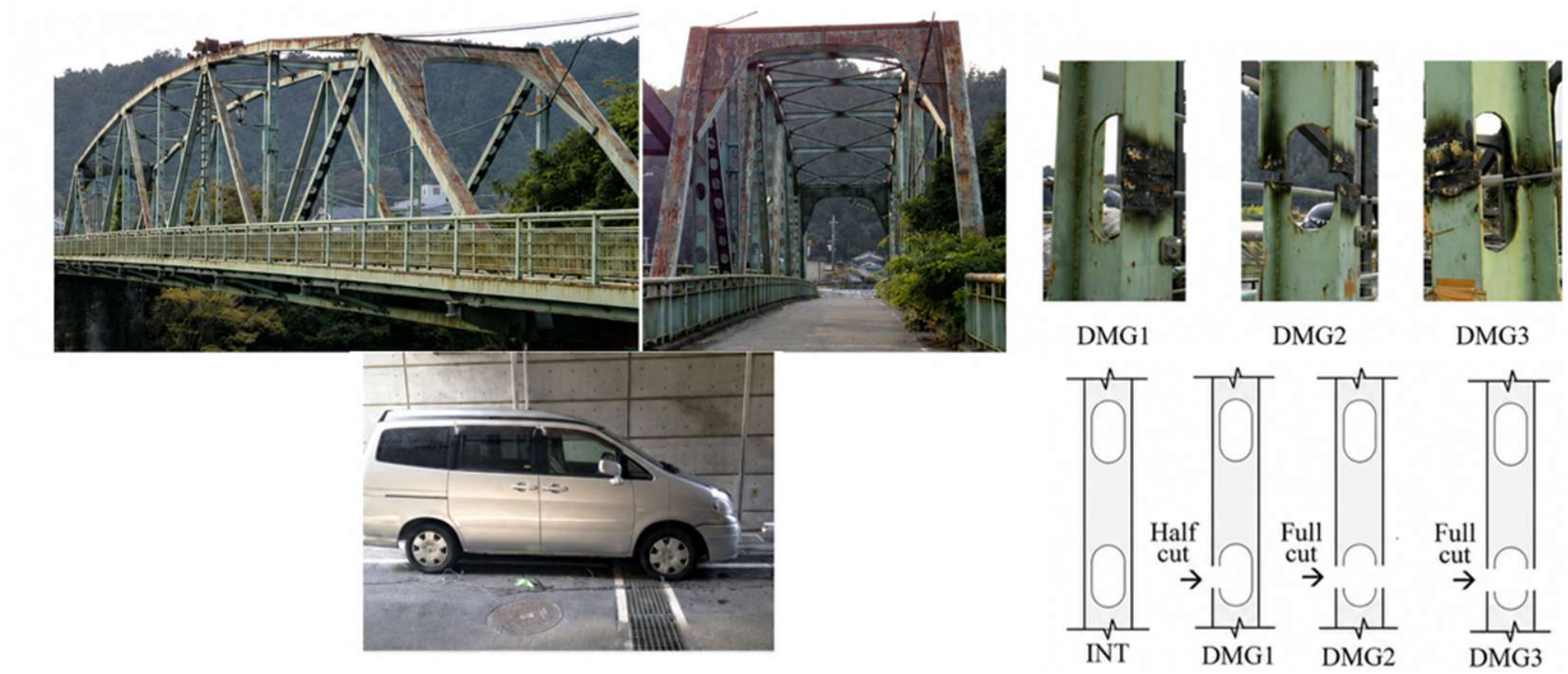

**Figure 2.** The Old Ada Bridge: the truss, an artificially damaged member, and the inspection vehicle [6].

Damage is represented as a reduction in Young's modulus of one entire member. The structural state is therefore a vector of 29 severities. This finite vector, rather than a spatially continuous damage field, distinguishes the truss problem from beam-type formulations in the literature [3]. The operator is a Fourier Neural Operator (FNO) [2], in which the learned mapping is parameterised spectrally and acts directly on the response's frequency content. In the forward configuration, the 29 severities are entered as constant input channels, and the operator returns the response measured at the vehicle's two axles over the entire crossing. Because damage alters the response far less than the crossing itself excites it, the operator is trained on the residual, the response difference between the damaged and healthy crossings. In the inverse configuration, the axle responses are the input, and the operator returns the damaged member and its severity through two output heads. The operator is trained on 1,160 simulated crossings, 40 per member, at severities between 0.10 and 0.60, of which 986 are used for training, 58 for validation, and 116 are withheld for testing. Figure 3 illustrates the forward configuration.

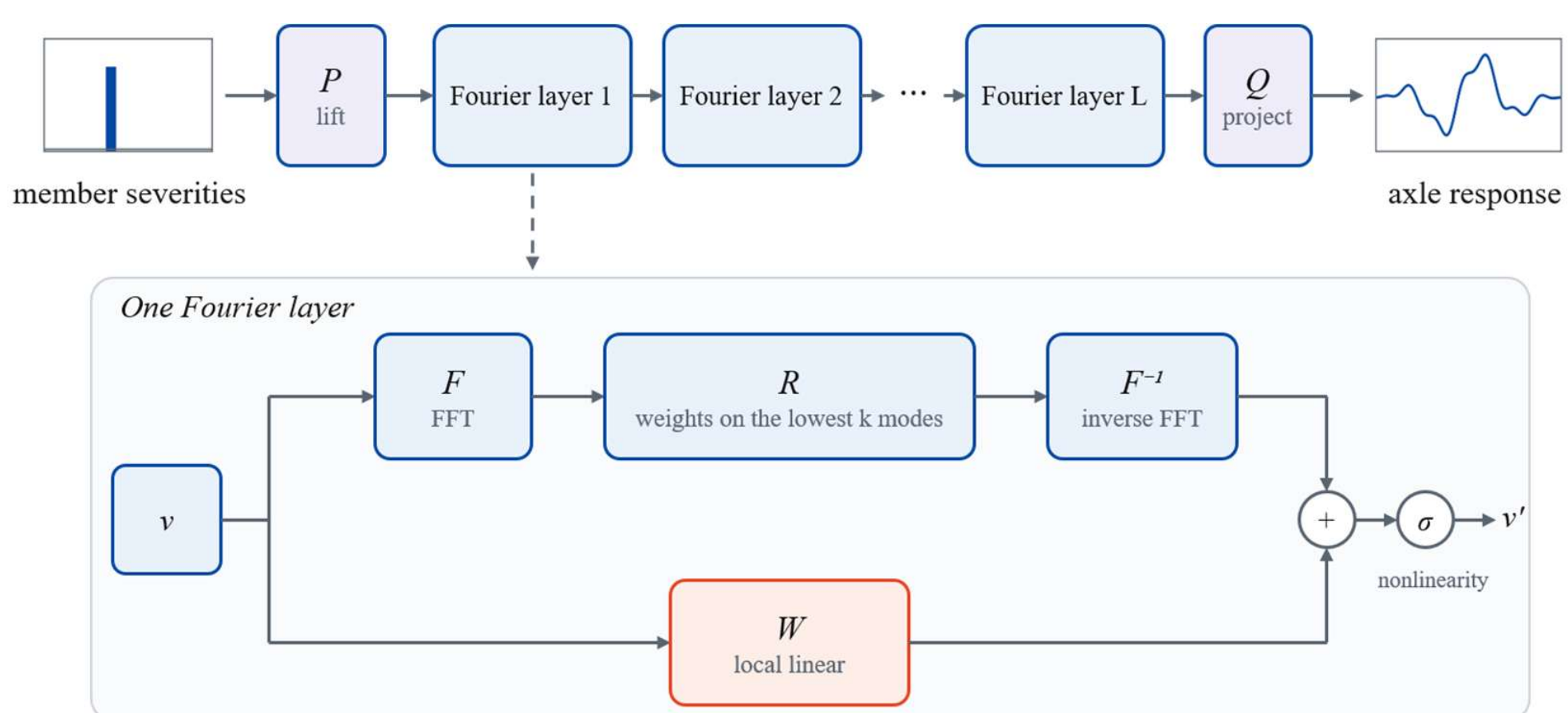

**Figure 3.** Concept of the neural operator for the forward surrogate model.

## 3. RESULTS

All results below are reported on the 116 withheld crossings. In the forward direction, the predicted vehicle response closely tracks the FE solution, as shown in Figure 4. The FNO requires only a small fraction of the computational cost of the FE solution. Its accuracy improved steadily as the training set grew, so the forward twin is presently limited by data rather than by its capacity. The remaining errors follow a regular pattern: accuracy reflects how strongly the damage manifests in the measured response.

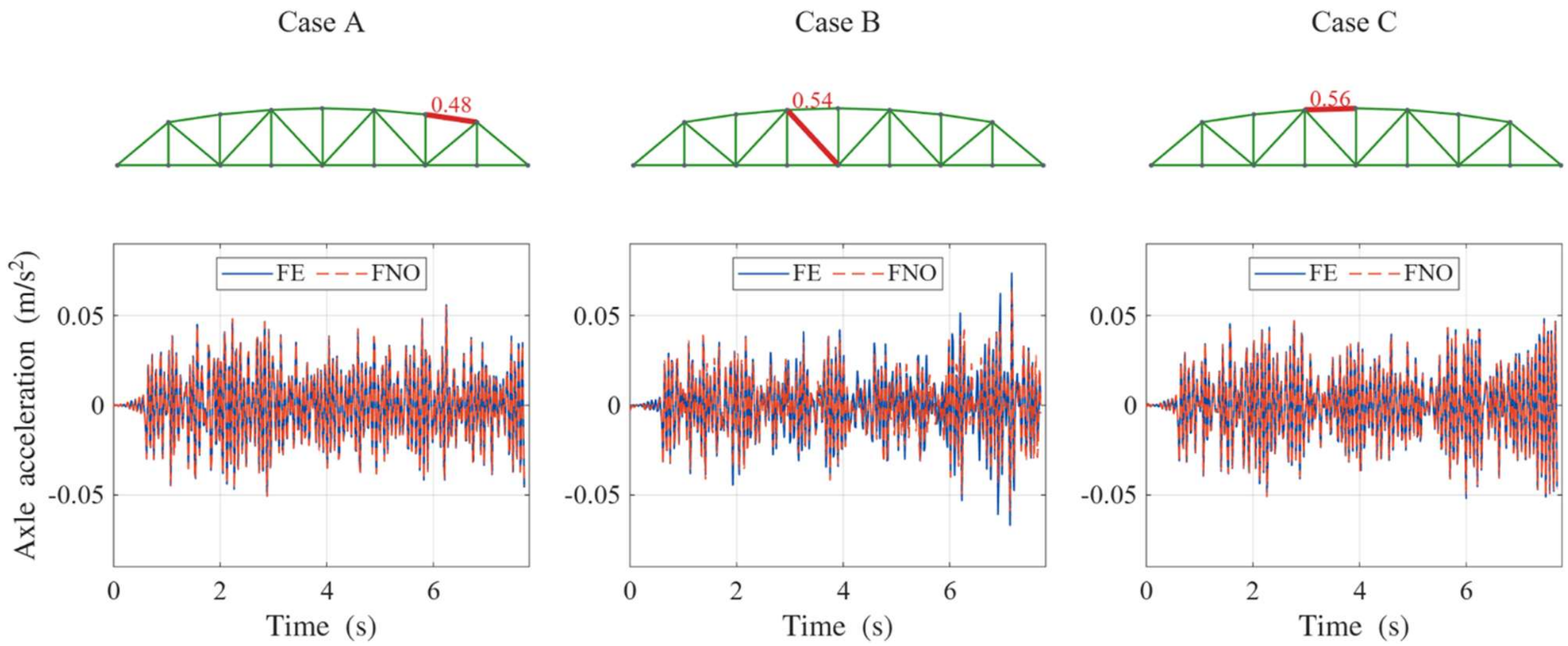

**Figure 4.** Forward results: 3 damage cases (damaged member in red) and the corresponding axle accelerations.

Figure 5 shows selected inverse results: Cases A and B are successful identifications, and Case C is a misidentification. Nonetheless, the FNO digital twin correctly identifies the damaged member in 102 of the 116 withheld crossings. The mean absolute error in the estimated severity is 0.03. The failure cases follow the same pattern: a misidentification selects a member immediately adjacent to the true one within the same chord. Even so, the estimated severity remains accurate. What remains is a question of distinguishability rather than observability: adjacent chord members alter the vehicle response in nearly identical ways, and the estimated damage then splits between the members concerned. Overall, the FNO approach works well with drive-by inspection. The vehicle is a moving sensor that passes over the entire span and responds to damage anywhere along the bridge, which enables the FNO digital twin to identify and localise damage.

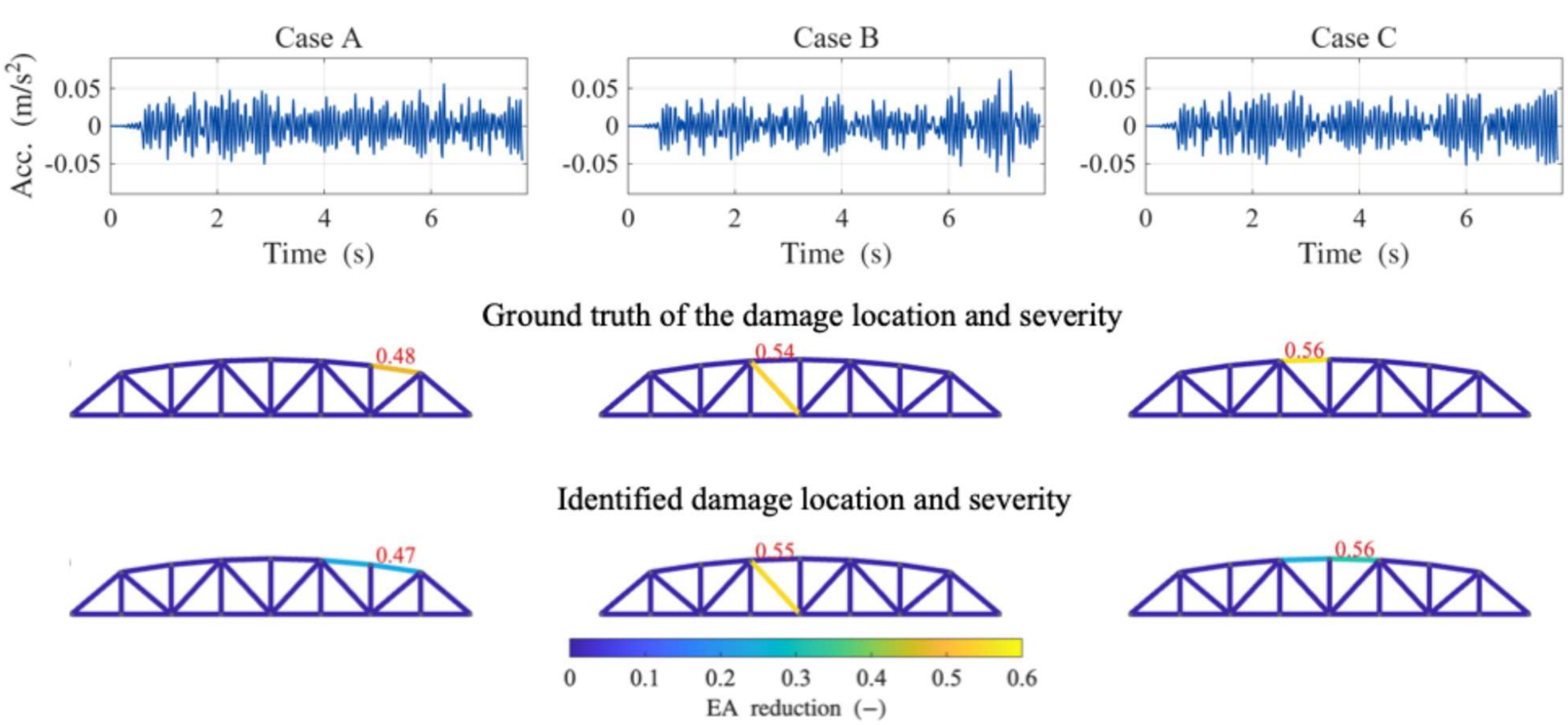

**Figure 5.** Inverse results: the recorded axle accelerations, the true damage, and the estimated damage (member probability times estimated severity).

## 4. CONCLUSIONS

A neural-operator digital twin of the Old Ada Bridge was trained on simulated vehicle crossings with member-level damage in the drive-by configuration: the crossing vehicle's response serves as the twin's output in the forward direction and as its input in the inverse direction. The main findings are drawn hereafter. First, the twin closely reproduces the vehicle response of the full FE simulation, several orders of magnitude faster, making continuous forward evaluation practical. Second, its inverse configuration identifies the damaged member correctly in 102 of the 116 crossings withheld from training and accurately recovers the severity. It is also found that the prediction accuracy depends on how strongly the damage affects the measured response. The moving vehicle is advantageous in this respect. Because it passes over the entire span, it captures the effect of damage in members near the supports, which a fixed sensor at a single location can barely see. The present work is a numerical case study, and future work will validate the twin against field data from the Old Ada Bridge [6].

## ACKNOWLEDGEMENTS

This work is supported by the Australia-Japan Innovation Fund (AJIF).